%% file: main.tex
\documentclass{article}
\PassOptionsToPackage{sort}{natbib}
\usepackage{iclr2026_conference,times}
\input{math_commands.tex}

\usepackage{amsmath,amssymb}
\usepackage{graphicx}
\usepackage{hyperref}
\usepackage{booktabs}
\usepackage{url}
\usepackage{placeins}

\title{Cross-View World Models}

\author{%
  \textbf{Rishabh Sharma}$^{1, 2, \ast}$ \qquad
  \textbf{Gijs Hogervorst}$^{3}$ \qquad
  \textbf{Wayne E. Mackey}$^{3}$ \\[0.3em]
  \textbf{David J.\ Heeger}$^{4,5}$ \qquad
  \textbf{Stefano Martiniani}$^{1,2,4,6,\ast}$ \\[0.8em]
  $^{1}$Simons Center for Computational Physical Chemistry, New York University \\
  $^{2}$Center for Soft Matter Research, Department of Physics, New York University \\
  $^{3}$Statespace Labs, Inc. \\
  $^{4}$Center for Neural Science, New York University \\
  $^{5}$Department of Psychology, New York University \\
  $^{6}$Courant Institute of Mathematical Sciences, New York University \\[0.5em]
  {\normalfont $^{\ast}$Corresponding authors: \texttt{rs10125@nyu.edu}, \texttt{sm7683@nyu.edu}}
}

\iclrfinalcopy
\begin{document}
\maketitle

\begin{abstract}
World models enable agents to plan by imagining future states, but existing approaches operate from a single viewpoint, typically egocentric, even when other perspectives would make planning easier; navigation, for instance, benefits from a bird's-eye view. We introduce Cross-View World Models (\textbf{XVWM}), trained with a cross-view prediction objective: given a sequence of frames from one viewpoint, predict the future state from the same or a different viewpoint after an action is taken. Enforcing cross-view consistency acts as geometric regularization: because the input and output views may share little or no visual overlap, to predict across viewpoints, the model must learn view-invariant representations of the environment's 3D structure. We train on synchronized multi-view gameplay data from Aimlabs, an aim-training platform providing precisely aligned multi-camera recordings with high-frequency action labels. The resulting model gives agents parallel imagination streams across viewpoints, enabling planning in whichever frame of reference best suits the task while executing from the egocentric view. Our results show that multi-view consistency provides a strong learning signal for spatially grounded representations. Finally, predicting the consequences of one's actions from another viewpoint may offer a foundation for perspective-taking in multi-agent settings.

\end{abstract}

\section{Introduction}

Good world models~\citep{ha2018worldmodels,lecun2022path} are faithful internal representations of an environment's dynamics, cheaper to query than acting in the real world. This allows agents to iterate on policy internally---that is, to \textit{plan}---before taking actions, learning from imagined actions and their outcomes. Having an internal world model is paramount for robust, grounded agents, especially those that must interact with the physical world.

Despite tremendous progress in action-conditioned video generation for world modeling~\citep{diamondWM,bar2024nwm,gameNgen,oasis2024}, current approaches predict from a single perspective, almost always egocentric. While such models may implicitly learn aspects of 3D structure, their training objective does not require it. The model can minimize its loss through patterns that hold only within that particular viewpoint.

Humans, too, see the world from an egocentric viewpoint. Yet we can shift to novel perspectives when we plan. For navigation, we can imagine a map of our environment and traverse it ``from above.'' Problems often become trivial after reframing them in the right coordinate system. Vision is a malleable mode of thinking, and perspective shifts are integral to it.

When world models serve as internal planning engines, the egocentric viewpoint is not always the most efficient frame of reference. Consider navigation: current world models~\citep{bar2024nwm} plan by generating ensembles of autoregressive rollouts, requiring thousands of diffusion passes even in known environments, only to check how close each trajectory came to the goal. This is an inefficient abstraction. Mammalian brains solved navigation differently, using specialized cells in the hippocampus that form a cognitive map and allow the animal to locate and track itself in space. Map-based navigation is more robust than the route-based planning that current world models perform.

\begin{figure}[t]

  \centering
  \includegraphics[width=1.0\linewidth]{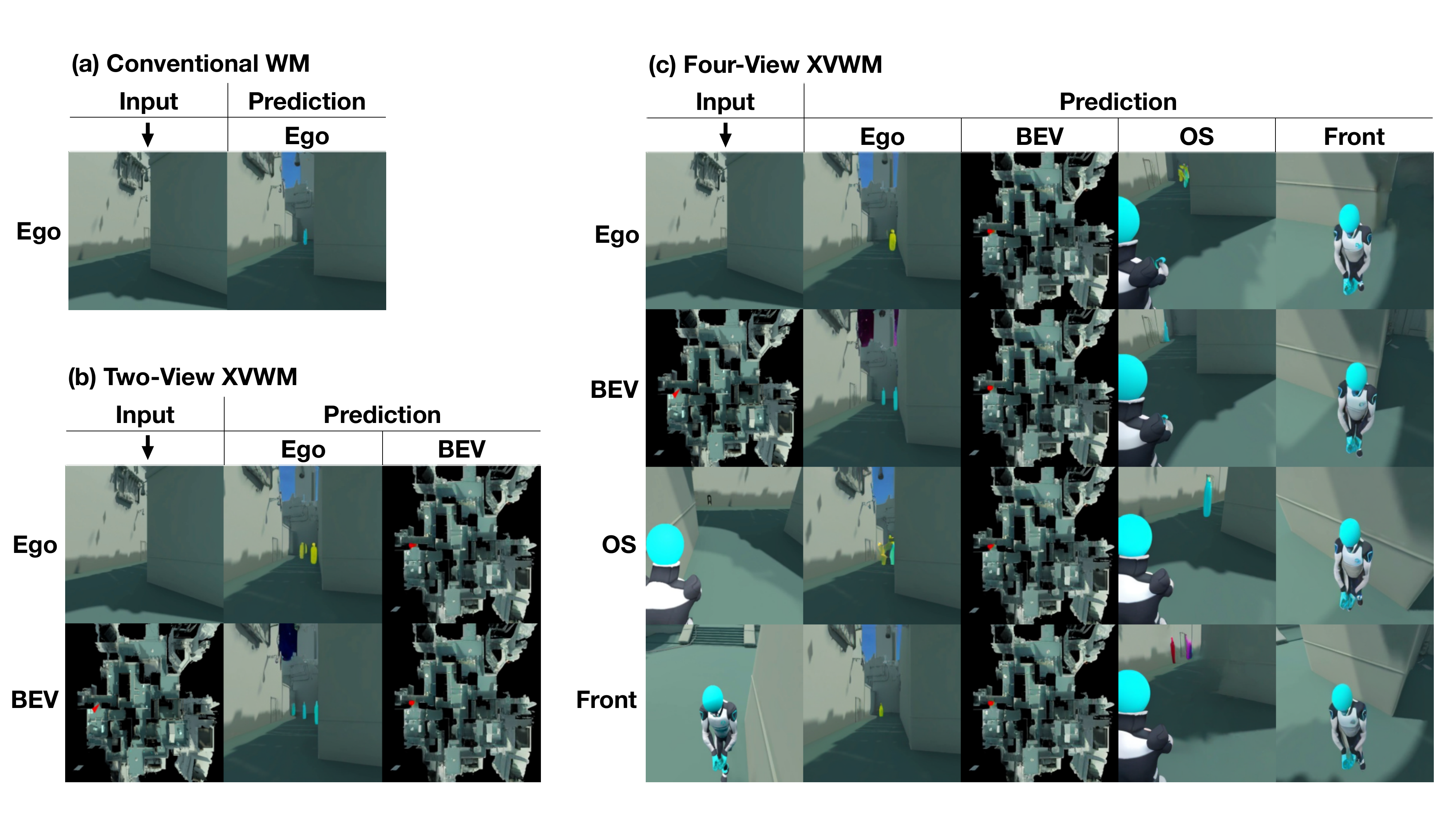}
   \caption{\textbf{Gaining multiple imagination streams.} 
(a)~A conventional world model predicts future states from the same 
viewpoint. (b)~An XVWM trained on egocentric and bird's-eye views 
learns bidirectional cross-view prediction: given either viewpoint as 
input, it can also imagine the future state from the other. These results yield a functional analog of biological cognitive maps: the model can locate and orient itself in its environment from visual cues alone. (c)~Training 
on four viewpoints yields any-to-any prediction across all $16$ 
input--output view pairs. Given context from a single viewpoint, XVWM 
can predict action-conditioned future states from all available 
perspectives. All predictions shown are 4 seconds into the future 
with identical state and action conditioning. Only the last of the 4 context frames is shown as input.}
  \label{fig:figure1}
\end{figure}

We introduce Cross-View World Models (\textbf{XVWM}), trained with a cross-view prediction objective: given frames from one viewpoint as context along with an action, predict the future from any viewpoint in a fixed set (Fig.~\ref{fig:figure1}). This produces multiple parallel and consistent imagination streams, so that given the same context and action, the model can predict the consequences of those actions from any desired viewpoint. See our model in action\footnote{\label{fn:XVWM_demo}\url{https://streamable.com/d62q7l}}. Enforcing cross-view consistency acts as geometric regularization. This is the first use of cross-view prediction as a self-supervised objective for training world model dynamics. Unlike joint multi-view prediction, where the model observes all viewpoints simultaneously and need only maintain consistency between them, cross-view prediction requires transforming from one perspective to another. The input and output views may share little or no visual overlap, so pixel-level patterns do not transfer. To predict accurately, the model must learn view-invariant representations of the environment's 3D structure, capturing the true physical state underlying all perspectives.

We train on synchronized multi-view gameplay data from Aimlabs, a first-person shooter aim-training platform providing precisely aligned multi-camera recordings with high-frequency action labels. When one of the views is a bird's-eye view (BEV) perspective with a marker indicating the agent's location and orientation, cross-view training yields spatial grounding, and the model learns to locate itself in its environment from visual cues alone. We also find that cross-view prediction provides positive transfer to same-view prediction, when the chosen views provide complementary information. For instance, egocentric and BEV are complementary, as they capture details at very different scales.

Beyond planning, perspective-taking has benefits in multi-agent environments. Both predators stalking prey and prey evading predators benefit from modeling what others can see. Reasoning about visibility and anticipating others' observations are plausible precursors to the theory of mind observed in higher primates~\citep{selman1971,YANG2015,Cohen1992}, and these are capacities that advanced AI agents will need to navigate the social world.

Our main contributions are as follows:
\begin{itemize}
\item We introduce cross-view prediction as a self-supervised objective for training world models, producing multiple parallel, switchable imagination streams from a single context-action pair.
\item We show that cross-view training with a bird's-eye view yields spatial grounding analogous to cognitive maps in mammals: from local visual cues alone, the model infers its global location and orientation, enabling map-based planning in known environments.
\item We demonstrate that cross-view training provides positive transfer to same-view prediction when trained with rich, complementary views.
\end{itemize}

\section{Related work}
\textbf{World models for planning.} World models learn environment dynamics to enable planning through imagination. The Dreamer family~\citep{hafner2019dream, hafner2020mastering, hafner2023mastering} uses recurrent state-space models to learn latent dynamics and has achieved strong results across diverse domains. More recently, approaches based on action-conditioned video generation have emerged. Genie~\citep{bruce2024genie} introduced generative interactive environments trained from unlabelled video, followed by models such as DIAMOND~\citep{diamondWM}, GameNGen~\citep{gameNgen}, and NWM~\citep{bar2024nwm}. While these models generate high-fidelity future frames, they predict from a single, typically egocentric, perspective. Our work removes this restriction, allowing a single model to imagine from any viewpoint in its training set. We build on the conditional diffusion transformer architecture of NWM~\citep{bar2024nwm} and adapt it for cross-view prediction.

\textbf{Multi-view representation learning.} Several works leverage multiple viewpoints to learn better representations. Contrastive Multiview Coding~\citep{tian2020contrastive} and related approaches~\citep{SimCLR,zhang2025,caron2021,grill2020} use a contrastive objective to learn embeddings that capture information shared across views. Time-Contrastive Networks~\citep{sermanet2018time} use synchronized video for self-supervised imitation. Computer vision approaches have explored cross-view synthesis~\citep{regmi2018cross, kulhanek2022viewformer}, though without action conditioning. Most relevant to our work is Multi-View Masked World Models (MV-MWM)~\citep{seo2023multi}, which trains a multi-view masked autoencoder to reconstruct randomly masked viewpoints, then learns a world model on frozen representations from the encoder. The key distinction is architectural: MV-MWM uses cross-view signals to train the encoder via reconstruction at the current timestep, while the dynamics model operates on frozen features without any cross-view objective. In XVWM, cross-view prediction trains the dynamics model directly: the model must predict the future state in a different viewpoint than the input. This forces the transition function itself—not just the encoder—to learn view-invariant representations of scene structure and dynamics.

\textbf{Multi-view 3D representations.} Neural Radiance Fields (NeRF)~\citep{nerf2020} and 3D Gaussian Splatting~\citep{kerbl3Dgaussians} learn to synthesize novel views from multi-view images. Recent works have integrated dynamics into these representations~\citep{lu2025gwm, zhang2024dynamic}, though they typically require explicit 3D supervision or incur high rendering costs. In autonomous driving, BEVFormer~\citep{li2022bevformer} and Lift-Splat-Shoot~\citep{philion2020lift} fuse multiple views into bird's-eye-view representations using explicit geometric projection, requiring known camera intrinsics and extrinsics. XVWM learns geometric relationships implicitly through the prediction objective, without camera calibration. In robotic manipulation, Ctrl-World~\citep{2025ctrlworldcontrollablegenerativeworld} takes all camera streams as input and generates all streams jointly, enforcing multi-view consistency through simultaneous prediction. In contrast, XVWM operates from a \textit{single} input view and generates any desired output view. This cross-view transformation task is strictly harder: the model cannot rely on correlations between input and output views, since they depict the scene from entirely different perspectives. To solve it, the model must internalize the 3D structure of the environment and re-render from the target viewpoint. This makes cross-view prediction a form of geometric regularization, forcing view-invariant representations in a way that joint multi-view prediction does not.

\textbf{Spatial representations in neuroscience.} We draw inspiration from the mammalian brain's ability to build rich spatial representations. Place cells and grid cells in the hippocampal-entorhinal system form cognitive maps that allow animals to locate and track themselves in space~\citep{hippocampus, Hafting2005}. We show that cross-view training with a bird's-eye view can yield analogous spatial grounding, where the model infers its global location and orientation from local visual cues alone.

\section{Dataset and model details}

\textbf{Dataset.} We use $4{,}186$ one-minute gameplay (first-person shooter) episodes collected at 60 frames per second and 
subsampled to 5 frames per second. Images are $224 \times 224$px. Each frame is paired with the action taken at that timestep. 
Every episode is recorded from four perspectives: egocentric\footnote{\url{https://streamable.com/nutp36}}, 
bird's-eye view (BEV)\footnote{\url{https://streamable.com/ahaxwo}}, over-the-shoulder (OS)\footnote{\url{https://streamable.com/bkecg2}}, and front-facing (Front)\footnote{\url{https://streamable.com/e09f2w}} (see Fig.~\ref{fig:figure1} for stills from each).
OS and Front follow the agent from a fixed distance, whereas BEV is a fixed overhead view that marks the agent's position and facing direction with a red elongated triangle. Owing to the tactical nature of the game, colored targets 
occasionally appear and change color upon being hit. The sky varies across episodes, ranging from bright daytime to 
nighttime and stylized themes such as space. Notably, the BEV provides no sky context, which we later exploit as a natural test of generalization. In this work, we 
condition on translational movement and change in yaw $(\Delta x, \Delta y, \Delta \phi)$, and do not aim to capture shooting dynamics. We use a 90:10 train-test split.

\textbf{Architecture:} Our method is architecture-agnostic: any action-conditioned video generation model can in principle be extended with viewpoint conditioning. We choose to build on the Conditional Diffusion Transformer (CDiT) from Navigation World Models (NWM)~\citep{bar2024nwm}, which was shown to outperform many of its predecessors. We add a learnable view embedding table that maps each view ID to a dense vector of the same dimension as the model's hidden state. This view embedding is added to the existing conditioning signal -- comprising the diffusion timestep, relative time, and action embeddings -- which is injected into each transformer block via adaptive layer normalization~\citep{peebles2023dit, xu2019layernorm}. 
The model takes $4$ consecutive frames from any single viewpoint as context (amounting to a 0.8\,s context window at 5 FPS). Following NWM, the target frame can lie in either the future or the past relative to the context: during training, the prediction time horizon extends to 64 frames (12.8\,s) into the future or into the past, while at test time it only predicts up to 25 frames (4\,s) into the future.

\textbf{Training and Compute:} We train three models:

\textit{Single-View baseline:} Trained exclusively on egocentric data with standard same-view prediction. This serves as our baseline.

\textit{Two-View XVWM (cognitive-map-inspired):} Trained on egocentric and bird's-eye views. At each training step, the input view is sampled uniformly at random. The output view is then sampled independently, with a cross-view probability of $0.5$: half the time the target matches the input view, half the time it differs. This yields all four input--output combinations (ego$\to$ego, ego$\to$BEV, BEV$\to$ego, BEV$\to$BEV) with equal probability $0.25$.

\textit{Four-View XVWM (generalist):} Trained on all four views. At each training step, the input view is sampled uniformly at random. With probability $0.5$, the output view matches the input (same-view prediction); otherwise, the output view is sampled uniformly from the remaining three views (cross-view prediction). This yields same-view pairs at $12.5\%$ each and cross-view pairs at approximately $4.17\%$ each.

The cross-view probability is a design choice which can be further optimized. Same-view prediction is necessary for autoregressive rollouts, but one can weight perspectives according to how informative they are or how much the downstream task depends on a particular viewpoint.

All three models are trained on $4\times$ H100/H200 GPUs with an effective batch size of 64 for 73 epochs, using the CDiT-L architecture.

\section{Results}

\begin{figure}[h]
  \centering
  \includegraphics[width=1.00\linewidth]{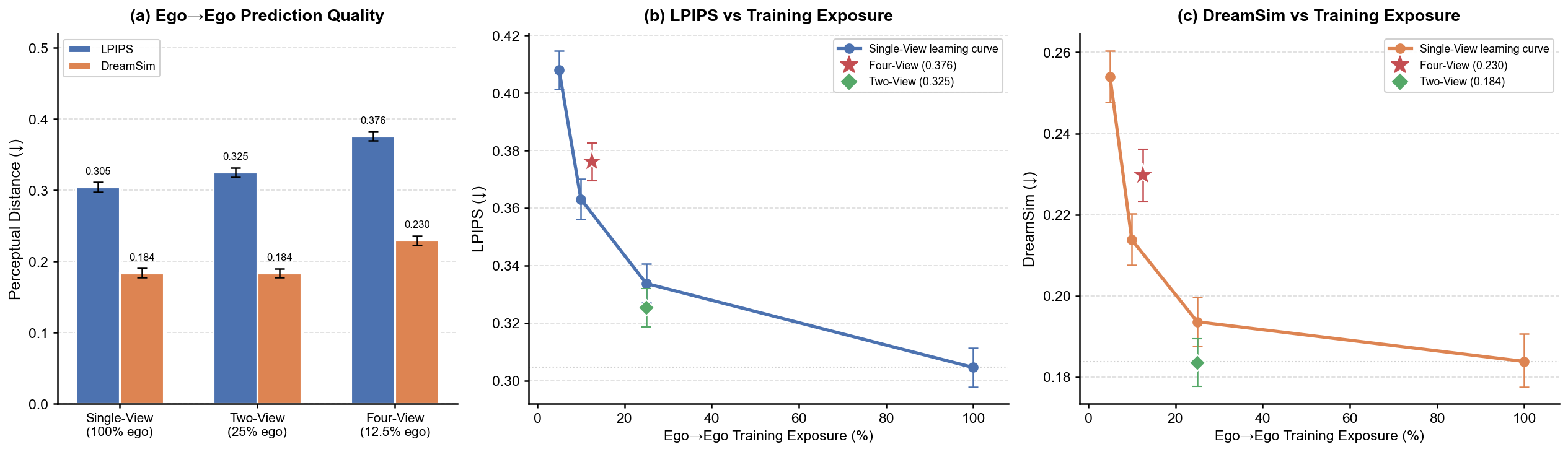}
\caption{\textbf{Same-view prediction quality and transfer.} 
\textbf{(a)} Ego$\to$ego perceptual similarity (LPIPS and DreamSim, 
lower is better) across the three models. The Single-View baseline 
sees $100\%$ egocentric pairs, the Two-View model $25\%$, and the 
Four-View model $12.5\%$. All models are trained with identical 
compute (same steps, epochs, and batch size). 
\textbf{(b, c)} Ego$\to$ego quality plotted against training 
exposure. Blue/orange curves show the Single-View model at 
intermediate checkpoints; starred and diamond markers show the fully 
trained Four-View and Two-View XVWMs at their effective exposures. 
Metrics are computed for a 4-second prediction target, averaged over 
$1{,}257$ test samples ($3$ predictions per test episode). Error bars 
show bootstrap $95\%$ confidence intervals.}
\label{fig:figure2}
\end{figure}

We evaluate XVWM along several dimensions: same-view and cross-view generation quality, perceptual similarity to ground truth, localization accuracy in the bird's-eye view, and spatial consistency of imagined trajectories. We also explore the model's ability to spawn anywhere in a known environment, reconstructing accurate egocentric scenes solely from the placement of a ${\sim}17$px marker in BEV.

Our results suggest that the models not only learn to transform between views, but also learn to transform across scales. Fast movements in the egocentric view must translate to slow movements of the marker on the BEV map, and vice versa. Similarly, tiny angular changes seen from above can translate to the entire field of view shifting drastically at eye level. We also investigate the importance of complementary, information-rich views for positive transfer, and show how redundant or visually impoverished views can degrade performance. Together, these experiments probe how cross-view training yields a coherent internal representation of 3D space.

Figure~\ref{fig:figure1} shows the capabilities of XVWMs at a glance: 
given an input view and action, the model predicts future states from 
multiple viewpoints, producing parallel imagination streams. Note the consistency across columns: each output depicts the same underlying scene from a different perspective, suggesting the model has learned a coherent representation of the 3D environment.

\subsection{Perceptual similarity and Transfer} 
We begin by systematically probing the same-view and cross-view 
generation quality for all three models, using LPIPS~\citep{lpips2018} 
and DreamSim~\citep{dreamsim2023} as perceptual metrics. We find that static elements---namely the fixed map in BEV, and the always-present 
agent body in OS and Front---saturate these scores to different 
extents, making comparison across views unreliable (see 
Appendix, Fig.~\ref{fig:appendix1}). We thus limit perceptual quality 
analysis to ego as the target view for both same-view and cross-view predictions. The LPIPS score for the Single-View model is comparable to that reported for NWM (CDiT-L)~\citep{bar2024nwm} trained on RECON, and serves as our baseline.

Starting with same-view predictions: Fig.~\ref{fig:figure2}(a) 
reveals that even after seeing dramatically fewer ego$\to$ego pairs during training, the Two-View XVWM matches the perceptual quality of the fully trained baseline on DreamSim, strongly suggesting positive transfer from cross-view to same-view prediction. The Four-View model, on the other hand, suffers some degradation. Given the dramatically fewer samples seen by both models, it is important to systematically explore transfer by normalizing for training exposure. 

We investigate this by comparing the multi-view models against the Single-View baseline at matched training exposure 
(Fig.~\ref{fig:figure2}(b,c)). The Two-View model (egocentric + BEV) consistently outperforms the exposure-matched baseline on both metrics, confirming positive transfer from cross-view training. We attribute this to the complementary information content of the two views: egocentric provides rich local context at eye level, while BEV provides a global map with precise agent location and orientation. Together, they offer more learning signal than either alone.

The Four-View model, which adds over-the-shoulder and front-facing views, performs consistently worse than the Two-View baseline despite similar training budgets. This may still be a worthwhile trade-off given the two additional imagination streams, but the degradation warrants explanation. We attribute it to differences in view informativeness. The over-the-shoulder view overlaps substantially with the egocentric perspective and tilts slightly toward the ground, providing limited additional signal. The front-facing view is more problematic: it faces the agent and angles downward, devoting much of the frame to the monotonic ground plane. In tactical scenarios where the agent moves along walls, this view collapses to a near-uniform surface, stripping away the environmental context needed to disambiguate one location from another.

These results suggest that information content of the views matters more than just view count. Implementing XVWMs with complementary views---such as egocentric and BEV---is key to positive transfer. Adding redundant or low-information views can dilute the learning signal and even degrade performance.

\begin{figure}[t]
  \centering
  \includegraphics[width=0.80\linewidth]{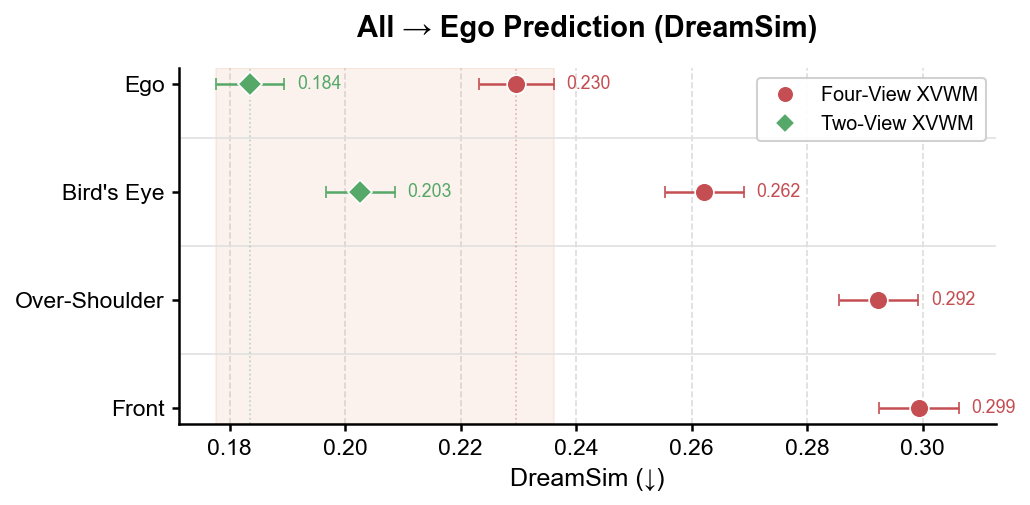}
\caption{\textbf{Cross-view prediction quality (all$\to$ego) within and across XVWM models.} The Two-View model, trained on 
complementary, high-signal views, outperforms the 
Four-View model. This is partly explained by the Two-View model's 
higher exposure ($25\%$ across all pairs) compared to the Four-View model ($12.5\%$ and $4.17\%$ for same-view and cross-view pairs respectively). However, even within the Four-View model, OS and Front perform considerably worse than BEV at similar exposure levels. Metrics are computed for a 4-second prediction target, averaged over $1{,}257$ test samples ($3$ predictions per test episode). Error bars depict bootstrap $95\%$ confidence intervals.}
\label{fig:figure3}
\end{figure}

Our analysis is consistent when considering cross-view performance. The Two-View model again decisively outperforms the Four-View model (Fig.~\ref{fig:figure3}), with over-the-shoulder and front being the primary underperformers. However, to properly normalize for training exposure, one would need base models trained on each view independently, analogous to what we did for same-view prediction. An analogous plot depicting LPIPS scores can be found in Fig.~\ref{fig:appendix2}, which follows the same trend.

\subsection{Localization and Trajectory Consistency} 

As discussed, perceptual metrics like LPIPS and DreamSim are of 
limited use for BEV evaluation: the static map dominates the score, saturating both metrics. To isolate the signal of interest, we instead quantify marker localization---the predicted position of the agent marker compared to ground truth. 
We measure the median distance in pixels between the centroid of the ground-truth marker and the predicted marker across 1,257 test samples, for all input views targeting BEV (Table~\ref{tab:table1}). For reference, the marker is ${\sim}17$px on the long side on $224\times224$px images, and all predictions reported in the table are for 4 seconds into the future. Across all models and input views, we observe sub-marker-size median error, indicating strong spatial grounding. The Two-View model definitively outperforms Four-View, consistent with both, having seen $2\times$ more BEV-targeted pairs during training, and being only exposed to high quality views, unlike the Four-View model. Success@$k$ denotes the percentage of predictions with error below $k$ pixels.

\begin{table}[t]
\centering
\caption{Localization accuracy in bird's-eye view (BEV). 
\\}
\label{tab:table1}
\begin{tabular}{lllccc}
\toprule
XVWM & Input View & Output View & Median Error (px) & Success @5px & Success @10px \\
\midrule
Two-View & Ego & Bird's Eye & 2.0 & 93.3\% & 98.6\% \\
Two-View & Bird's Eye & Bird's Eye & 1.7 & 96.7\% & 99.3\% \\
\midrule
Four-View & Ego & Bird's Eye & 2.9 & 77.9\% & 95.2\% \\
Four-View & Bird's Eye & Bird's Eye & 2.6 & 81.8\% & 95.0\% \\
Four-View & Overshoulder & Bird's Eye & 3.7 & 64.4\% & 88.0\% \\
Four-View & Front & Bird's Eye & 3.8 & 62.7\% & 86.4\% \\
\bottomrule
\end{tabular}
\end{table}

\begin{figure}[ht]
  \centering
  \includegraphics[width=1.00\linewidth]{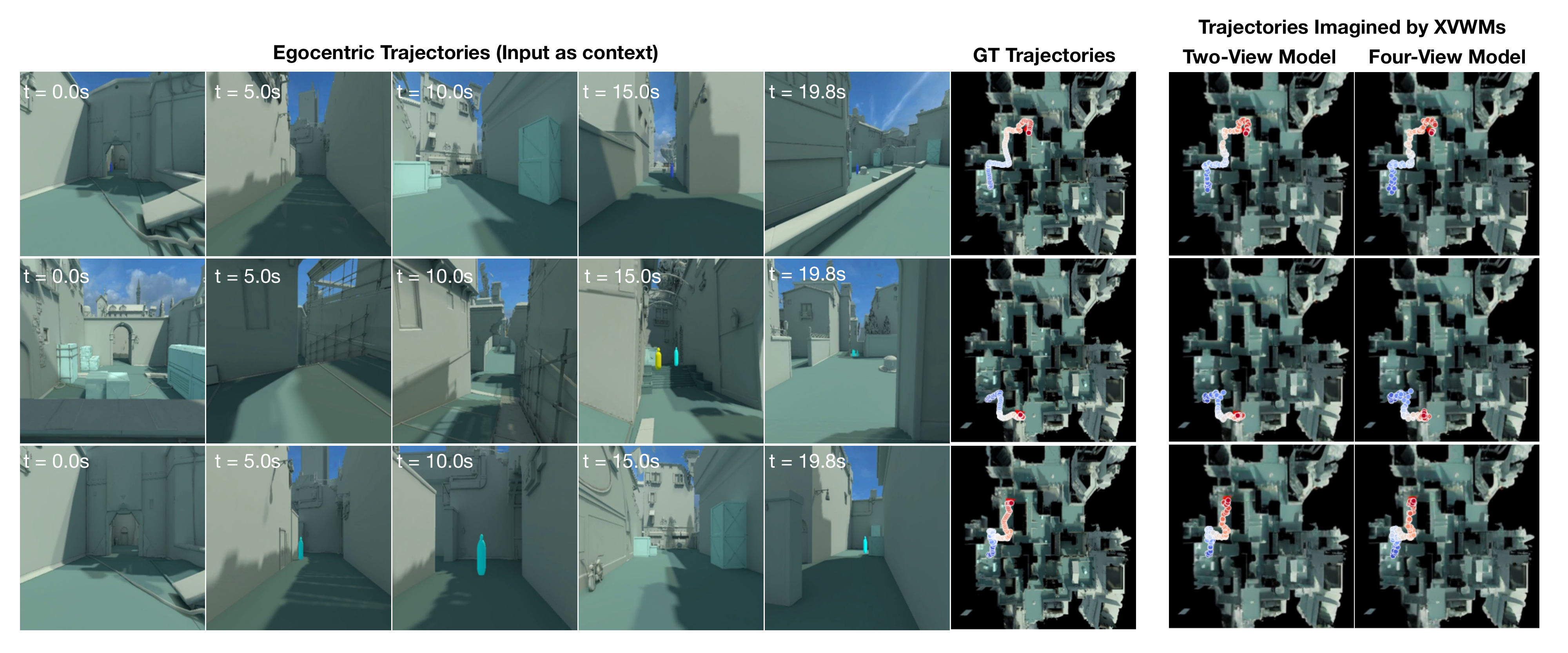}
  \caption{\textbf{XVWM's internal geolocalization.} Localization in an environment is but one component of the mammalian cognitive map. Localization alone is insufficient for navigation, one also need to encode consistent spatial ordering. Here, we probe this property by following three egocentric trajectories and imagining the corresponding path from BEV. Comparing to ground-truth trajectories, we observe that movements in the egocentric view translate to consistent movements on the BEV map. The blue-to-red gradient denotes time increasing from $0.0$s to $20.0$s. At each point, the model predicts the next frame, i.e it predicts $0.2$s into the future. The model possesses an ``internal GPS'' that allows it to locate and orient itself in its environment from visual cues alone.}
  \label{fig:figure4}
\end{figure}
Going beyond localization, a useful cognitive map must also capture spatial relationships between locations. Localization alone, even at sub-marker accuracy, is not sufficient. Fig.~\ref{fig:figure4} compares three egocentric trajectories taken by the agent against the BEV trajectories imagined by the model. The agreement with ground truth is striking, confirming that the model maintains a functioning internal ``GPS.'' This suggests a path toward training similar systems for autonomous vehicles and smart glasses, which already collect egocentric video and operate within mapped environments—though achieving the precise synchronization of our setup would require high-fidelity localization.
Example 
videos\footnote{\url{https://streamable.com/t1o1ym}}\textsuperscript{,}\footnote{\url{https://streamable.com/llv6z5}} 
show the agent following an egocentric path while imagining the 
corresponding path from BEV. Egocentric movements transfer 
consistently to corresponding imagined movements on the map. 

Accurately detecting orientation for a marker of this size is inherently noisy; however, the orientational accuracy can be gauged from the demo video\textsuperscript{\ref{fn:XVWM_demo}}. Orientation plays an even larger role in the inverse task of spawning, which we discuss in the next section.

\subsection{Spawn Anywhere}

\begin{figure}[h]
  \centering
  \includegraphics[width=1.00\linewidth]{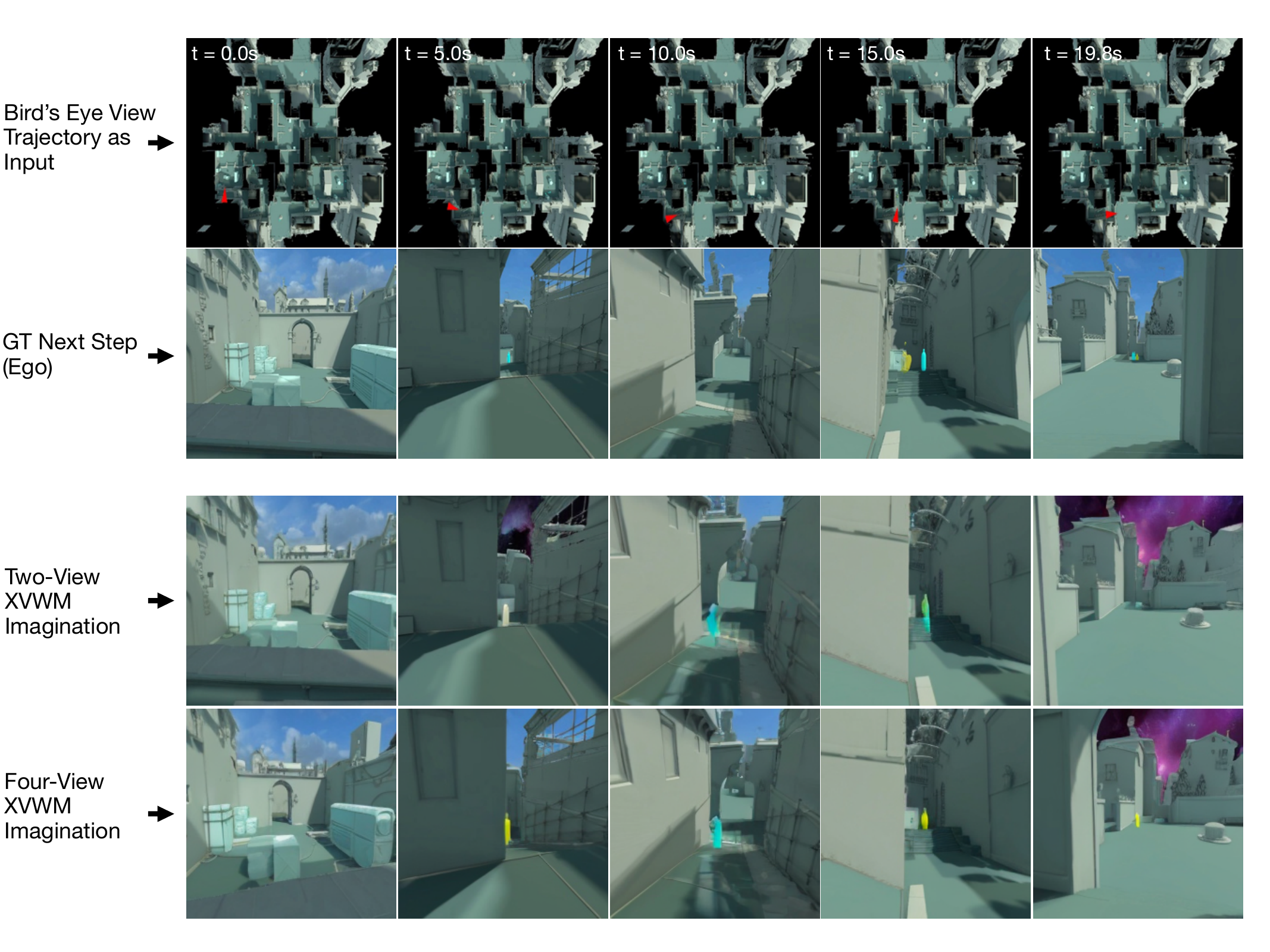}
  \caption{\textbf{Spawn anywhere in a known environment.} A dramatic consequence of the emergent bi-directionality from cross-view training is the ability to spawn at arbitrary locations. Here, we take a Bird's-eye view trajectory as input (top row) and predict the action-conditioned next step from an egocentric viewpoint. Both models imagine detailed egocentric views along the trajectory, decoding all the necessary information from the location and orientation of the tiny ${\sim}17$px marker alone. This suggests strong emergent understanding of the 3D environment. Notably, the sky varies stochastically when input context is BEV, as expected: the training environment included multiple sky layouts, and BEV provides no information about which sky should appear. This suggests the model has learned spatial structure rather than memorizing input-output pairs. At each instant, the model predicts the next frame, that is $0.2$s into the future.}
  \label{fig:figure5}
\end{figure}

Perhaps the most striking evidence that cross-view training yields spatial understanding lies not in localization (ego$\to$BEV), but in its inverse. Given any marker position on the map and any action, the model generates how the 3D scene will appear from that vantage point at the next instant: a rich visual prediction from a ${\sim}17$px marker. Even a slight change in marker orientation completely rotates the field of view in the egocentric frame, forcing the model to attend closely to both the marker's position and heading. The model must also predict an \textit{action-conditioned future}, precisely calibrating the slow marker dynamics observed in the BEV context to the much faster visual changes in egocentric. This is only possible if the model has internalized the spatial structure of the environment. Fig.~\ref{fig:figure5} demonstrates this capability.

BEV$\to$ego is a considerably harder task than ego$\to$BEV. In the latter, the model could in principle predict a static map while ignoring the marker altogether, though as we have shown, it does not. In the BEV$\to$ego direction, no such shortcut exists: the vast difference in information density between input and output, combined with the fact that small marker movements translate to dramatic viewpoint shifts, forces the model to attend to the marker and build correspondingly rich spatial representations. The tiny marker acts as an information bottleneck, and the asymmetry between sparse input and detailed output can only be bridged by genuine 3D understanding.

\FloatBarrier

\section{Conclusion and Future Directions}

Cross-view prediction is a surprisingly effective self-supervised 
signal for world models. Training on rich, complementary views 
transfers to better performance even on same-view prediction tasks. 
That is, the geometric constraints of cross-view prediction force 
view-invariant understanding that benefits all predictions, not just 
cross-view ones.

Beyond maintained quality, cross-view training yields capabilities unavailable to same-view models: sub-marker localization, trajectory consistency over 20-second rollouts, and the ability to generate detailed egocentric scenes from a $\sim$17px map marker. This last capability reflects a striking information asymmetry: the marker specifies only position and orientation, yet the model produces a rich 3D scene, which is only possible if it has internalized the spatial structure of the environment.

These results bring world models closer to the biological cognitive maps that inspired this work, but the analogy remains incomplete. Autoregressive world models have no persistent state beyond their context window and cannot update their internal representation as they explore. Biological cognitive maps are built incrementally through experience; ours must be trained offline on a fixed environment. Integrating persistent, updatable memory into cross-view world models is a natural next step.

Finally, the ability to imagine how the world appears from a viewpoint one does not occupy may have implications beyond planning. Reasoning about others' visual perspectives is a prerequisite for social intelligence. Whether cross-view world models can serve as a foundation for perspective-taking in multi-agent settings remains an open question.

\subsubsection*{Acknowledgments}
This work was supported by a grant from the Simons Foundation [MPS-T-MPS-00839534, MET] (RS, SM) and by the National Institute of Health under award number R01MH137669 (DJH, SM). RS, DJH, and SM gratefully acknowledge use of the research computing resources of the Empire AI Consortium, Inc., with support from Empire State Development of the State of New York, the Simons Foundation, and the Secunda Family Foundation~\citep{EmpireAI_2025}; as well as of additional computational resources and consultation support provided by the IT High Performance Computing at New York University.

\bibliographystyle{iclr2026_conference}
\bibliography{references}

\newpage
\appendix
\section{Supporting Results}
\begin{figure}[htpb]
  \centering
  \includegraphics[width=0.80\linewidth]{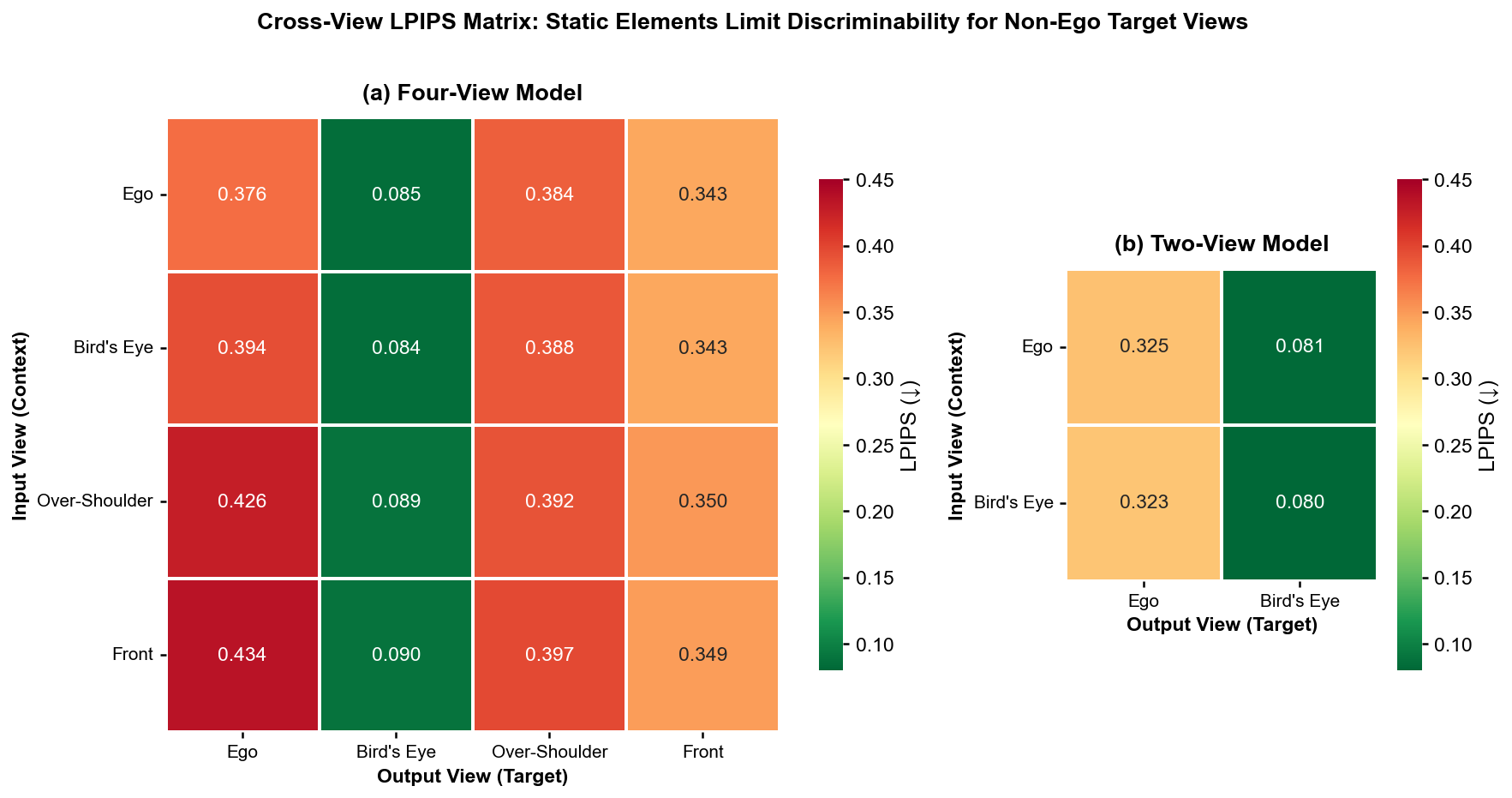}
\caption{\textbf{Cross-view LPIPS matrix.} The bird's-eye column 
shows uniformly low LPIPS regardless of input view, as the static 
background map dominates the score rather than meaningful agent 
localization. Similarly, over-the-shoulder and front columns contain static elements (the agent's body), reducing metric sensitivity. We therefore reserve perceptual metrics for ego-targeted predictions (both same-view and cross-view), where static elements do not dominate. BEV prediction quality is instead quantified using arrow position accuracy, trajectory consistency over time, and bidirectional spawn tests (map$\to$ego). Metrics are computed for a 4-second prediction target, averaged over $1{,}257$ test samples.}
\label{fig:lpips_matrix}
\label{fig:appendix1}
\end{figure}

\begin{figure}[htpb]
  \centering
  \includegraphics[width=0.80\linewidth]{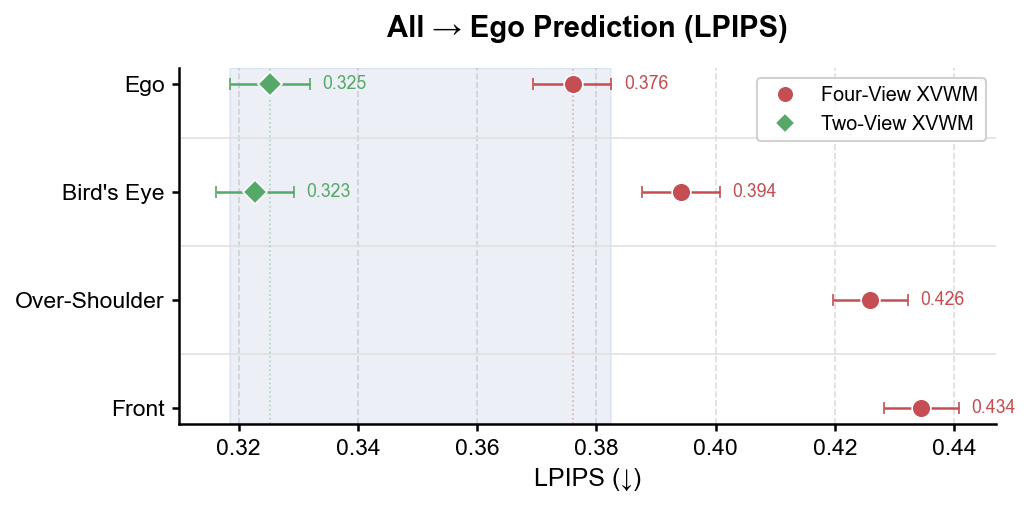}
\caption{\textbf{Cross-view prediction quality (all$\to$ego), LPIPS.} Same analysis as Figure~\ref{fig:figure3} using LPIPS instead of DreamSim. The same trend holds: the Two-View model outperforms the Four-View model, with OS and Front as the primary underperformers.}
\label{fig:appendix2}
\end{figure}

\end{document}

%% file: math_commands.tex
\usepackage{amsmath,amsfonts,bm}

\def\eqref#1{equation~\ref{#1}}

\def\1{\bm{1}}

\DeclareMathAlphabet{\mathsfit}{\encodingdefault}{\sfdefault}{m}{sl}
\SetMathAlphabet{\mathsfit}{bold}{\encodingdefault}{\sfdefault}{bx}{n}

